# Cooperative Automated worm Response and Detection ImmuNe ALgorithm(CARDINAL) inspired by T-cell Immunity and Tolerance

Jungwon Kim*, William O. Wilson†, Uwe Aickelin† and Julie McLeod‡

*Department of Computer Science, University College London, UK
j.kim@cs.ucl.ac.uk
†School of Computer Science, University of Nottingham, UK
wow,uxa@cs.nott.ac.uk
‡Faculty of Applied Science, University of the West England, UK
julie.mcleod@uwe.ac.uk

**Abstract.** The role of T-cells within the immune system is to confirm and assess anomalous situations and then either respond to or tolerate the source of the effect. To illustrate how these mechanisms can be harnessed to solve real-world problems, we present the blueprint of a T-cell inspired algorithm for computer security worm detection. We show how the three central T-cell processes, namely T-cell maturation, differentiation and proliferation, naturally map into this domain and further illustrate how such an algorithm fits into a complete immune inspired computer security system and framework.

## 1 Introduction

Self-propogating malicious code represents a significant threat in recent times as the ability of these programs to spread and infect systems has increased dramatically. The recent SQL Slammer worm infected more than 90% of vulnerable hosts on the Internet within 10 minutes [10], and at its peak the Code-Red worm infected over 2,000 hosts every minute [11]. Under such a constantly hostile environment, the traditional manual patching approach to protecting systems is clearly not effective.

An alternative solution to this problem is to have an automated detection and response system which could identify malicious self propogation and stop the spread of the worm as early as possible. Current automated detection and response systems involve such actions as blocking unsecure ports, dropping potentially threatening packets, and eliminating emails carrying malicious codes, breaking communication between infected and non-infected hosts to slow down worm propagation and minimise potential damage [12]. This appears to be a simple and obvious solution, however there are a number of significant hurdles to overcome in order to employ such automated responders. The most noteworthy obstacle is the high false positive error problem [16]. If an automated responder disconnects communication between two hosts based on a false positive result, the effect of this inappropriate disconnection could be as bad, if not worse than, the damage caused by the worm itself.

The objective of this paper is to propose a solution to this problem by taking inspiration from the Human Immune System (HIS). Previous research into computer security in the context of Artificial Immune Systems (AIS) has been focused on detecting unknown intrusions [2] [8]; detecting anomalous events such as abnormal network traffic patterns or abnormal sequences of system calls. However the reliability of these systems to handle non trivial problems is still in question as they have not yet passed tests to indicate that low false positives are achievable in a real environment [2] [8].

Instead of developing these existing AIS, we propose a novel AIS model that adopts numerous mechanisms inspired from the differentiation states of T cells. These differentiation states can be grouped into particular status subsets which can be used to classify the types of T cell. From these classifications, the various roles of the diverse T cell types can be seen in terms of their contribution to the unique aspects of overall immunity and tolerance within the HIS. In this paper we carefully study the significant properties and physiological mechanisms of each T cell subset, with regard to the way they influence the interaction of immunity and tolerance. This study allows us to design a new AIS model, CARDINAL(Cooperative Automated worm Response and Detection ImmuNe ALgorithm) which has the potential to operate as a cooperative automated worm detection and response system. The paper starts by addressing the research issues associated with such a system. Section 3 introduces the different differentiation states of T cells within the HIS. Section 4 presents a novel cooperative automated worm detection and response system which adopts CARDINAL and finally the paper concludes with details of future work planned.

## 2 Cooperative Automated Worm Detection and Responses

In order to detect the presence of a novel worm virus various automated anomaly detection and response based systems have been developed [12]. These systems trigger automated responses when they observe such things as abnormal rates of outbound connections, emails sent, or port scanning, etc. In order to improve the false positive error rate made by local anomaly detectors, an alternative cooperative strategy has also been suggested [3] [13] [14]. The motivation behind this approach is that additional information on the infectious status of the worm, and the responding states of other peer hosts, would help local responders make better decisions by taking into account the collective evidence on an attack's severity and certainty, and an infection growth rate. Indeed, some work has already reported that such a suggestion reduces false positive errors [16].

However, there are some significant issues to be tackled in order to make a cooperative strategy truly effective. Firstly, information shared between peer hosts should be lightweight, as the transfer of unnecessary and excessive information can create the potential for self denial-of-service attacks [3] [13]. Secondly response mechanisms should be robust against inaccurate information passed amoung hosts [3]. If the reaction to a false positive error is isolated to a single host, the impact is minimal. However because of the cooperative nature of the system, this inappropriate response could be disseminated to the rest of the network, causing other hosts to react in a similar fashion and exascerbate the problem. Thus, a cooperative system needs to localise the negative impact of such errors, and this could be done by constantly redefining the range of information to be shared in terms of an estimated certainty of detection results. In order to address these issues, we identify the following to be studied:

- **Optimise the number of peer hosts polled**: the CARDINAL system needs to determine which peer hosts are able to share information, and how many peer hosts should be selected to share that information. These decisions are directly aimed at preventing a possible break of self denial-of-service attack. Determining the set of peer hosts is done by identifying all the possible peer hosts that can be directly contacted and thus infected by a given host. However, the number of all possible peer hosts may be unnecessarily large as information shared by a smaller number of peer hosts might be sufficient to mitigate and stop worm propagation. An optimal number of peer hosts is desirable to mitigate the propogation of a worm to a sufficient degree whilst minimising the number of resources that are required to achieve that objective. The determination of the size of this optimal set of peer hosts would be influenced by factors such as the severity of the worm's threat, the certainty of attack detection, and the growth rate of the infection. The more severe an attack, the more certain we are of it being detected, or the faster is its propagation, then the larger the peer set needs to be so information can be shared by more peer hosts to counterattack the worm successfully.
- **Types of system responses should be determined by attack severity and certainty**: in order to reduce the negative effects of false detection results, CARDINAL selects its response to the threat depending on the certainty of an attack being detected and the severity of that attack. CARDINAL would respond to severe and certain attacks with strong actions, such as blocking ports showing anomalous outbound connection patterns, eliminating emails appearing to carry worms, or dropping hostile network packets containing attack signatures. Alternatively, when presented with relatively uncertain or less severe attacks, CARDINAL would take less severe action, such as logging the potential situation for an administrator or limiting the network connection rates.
- **For performing adequate magnitudes of responses, both local and peer information needs to be taken into account**: the severity and certainty of attacks should not be statically measured. A worm detected at a local host, at a given time, might appear to be relatively less severe, however if CARDINAL later observes that the number of peer hosts infected by the worm greatly increases within a short time frame, responses to this worm should be upregulated in terms of detection certainty and attack severity. The total number of infected peer hosts could be estimated based on the collective information passed between the peer set. Alternatively, when a severe attack is detected by a particular host, which disseminates this information to the remaining designated peer hosts, those hosts do not necessarily have to take the same corrective action as the original host. If the infectious symptoms are not shown at the peer hosts receiving this information, and the total number of infected peer hosts does not increase quickly, the peer host can change its response from a very strong reaction to a weaker one. In turn this host would decrease the number of other peer hosts to which it sends its detection and response information, curtailing the response to the worm and returing the system to a stable state. Considering these factors together, we see CARDINAL will determine the apppropriate number of hosts to be polled and the degree of response to a worm according to the severity and certainty of attacks, which are dynamically measured based on both local and peer information.

| CARDINAL | HIS |
| --- | --- |
| Optimise the number of peer hosts polled | Dynamically adjust the proliferation rate for each effector T cell |
| Types of system responses should be determined by attack severity and certainty | Differentiate appropriate types of effector T cells depending on interaction with cytokines and other molecules during the maturation proccess |
| For performing adequate magnitudes of responses, both local and peer information needs to be taken into account | T cell effector function is amplified and suppressed via interaction among different types of effector T cells |

Table 1. Mapping between CARDINAL and HIS

We believe that several mechanisms constituting T cell immunity and tolerance of the HIS could provide insight into intelligent approaches to implementing the previous three properties. Table 1 shows these three specific properties of T cells in the HIS, which were used to design CARDINAL. Section 4 discusses the details of these properties together with the proposed model of CARDINAL. Before this discussion, section 3 briefly reviews the various differentiation states of T cells and how they contribute to the HIS in balancing immunity and tolerance.

## 3 T-cell Immunity and Tolerance of HIS

The immune response is an incredibly complex process that one can argue begins with the dendritic cell (DC). DC's are a class of antigen presenting cell that migrate to tissue in order to ingest antigen or protein fragments. Whilst ingesting the antigen, DC's are also receptive to molecules in the environment that may be associated with the circumstances of that antigen's existence. These molecules are identified as a form of danger signal [9]. Once the antigen has been ingested in the tissue, the DC's travel back to the lymph nodes where they present the antigen peptides to naive or memory T cells via their MHC molecules, this allows a T cell to be able to identify that antigen. In addition, the DC will interpret the molecules it experienced during the ingestion process, and release particular cytokines[1] to influence the differentiation of the T cell it is presenting antigen to. In this way, the DC drives the T cell to react to the antigen in an appropriate manner and as such the DC can be seen as the interpretative brain behind the immune response. Given we now know what drives the T cell differentiation process, we turn to look at the different T cell differentiation stages. Much of this information has been taken from [5] [7] and reference to that work should be made if further detail is required.

### 3.1  Naive T cells

Naive T cells are T cells that have survived the negative and positive selection processes within the thymus, and have migrated to continuously circulate between the blood and lymphoid organs as they await antigen presentation by DC's. The important fact is that naive T cells have not experienced antigen and they do not as yet exhibit effector function.

### 3.2  Activated T cells

Naive T cells reach an activated state when the T cell receptor (TCR) on the surface of the naive T cell successfully binds to the antigen peptide-MHC molecules on the surface of the DC, and co-stimulatory molecules are sufficiently upregulated on the surface of the DC to reflect the potential danger signal. The degree of signaling from the DC influences the degree of activation of the T cells. T cells that receive high signal strengths adopt the potential for effector function and gain the ability to migrate from their current location in the lymph node to the periphery. These activated T cells gain the ability to proliferate and their clones will begin to differentiate into either helper T cells or cytotoxic T cells. These cells will finally reach effector status when they interact with a second antigen source. T cells that receive excessive levels of signalling die through a process of activation induced cell death (AICD) to prevent an excessive immune response taking place.

### 3.3  Helper T cells (Th)

Naive T cells express either CD4 or CD8 co-receptor molecules on their surface, so called as they are clustered with the TCR and bind to the MHC molecules

---

[1] Cytokines are chemical messengers within the HIS [5]. They are proteins produced by virtually all cells in the HIS and they play an important role in regulating the development of effector immune cells

presented on the DC. Naive T cells expressing CD4 differentiate into Th cells after activation. When they achieve effector status, through further antigenic stimulation, Th cells can develop into either Th1 or Th2 cells. The divergence between Th1 and Th2 is driven by the cytokines released from the DC when the T cell is first activated. Th1 and Th2 cells have different functionality as Th1 cells release cytokines that activate cytotoxic T cells whilst Th2 cells release cytokines that activate B cells.

In addition, a cross regulation mechanism exists between Th1 and Th2 cells. Cytokines released by Th1 cells directly impede the proliferation of Th2 cells, whilst Th1 cytokines downregulate the production of the cytokine IL-12 in DC's which in turn downregulates the proliferation of Th2 cells. This feedback mechanism leads to an immune response dominated by the particular Th cell subtype that is primarily stimulated, ensuring the more suitable immune response is initiated to resolve the current threat.

### 3.4 Cytotoxic T cells (CTL)

Naive cells that express the CD8 molecule on their surface are predestined to become CTL cells after activation. If the DC's themselves do not express sufficient co-stimulatory molecules to cause activation, then DC's can be induced to upregulate those signals by Th1 cells who also bind to the DC. Activated CTL's will undergo proliferation and migrate to inflamed peripheral tissues. When they receive stimulation from subsequent antigen, they will reach an effector status and develop the ability to produce antiviral cytokines and cytotoxic molecules, which when released will kill infected host cells that exhibit the antigen trace identified by the CTL. A CTL can bind to, and therefore kill, more than one infected cell at a time.

Current theories disagree as to whether, after reaching an effector state and carrying out their helper or killer function, CTL and Th cells either die as they have reached a terminally differentiated state or whether some proportion of the CTL / Th effector cell population differentiate into longer lived memory cells to facilitate a suitable secondary response.

### 3.5 Summary of T cell states

From the above sections, we can see that given the presentation of antigen by an APC and the existence of sufficient signals that indicate the presence of danger, a naive cell will become activated, will proliferate and differentiate into effector cells which can take on numerous alternative states. Depending on the co-receptors expressed on the effector T cell surface, these cells will either differentiate into Th or CTL cells. CTL cells lead the immune response by eliminating antigenic threats. Th cells provide assistance to this protective process but also provide regulation via a comprehensive feedback mechanism to ensure stabilisation. Naive cells that do not receive sufficient danger signals do not become activated and so the system becomes tolerant to such antigen strains. All these cells interact in a competitive environment that results in tolerance and immunity within the system.

## 4 Cooperative Automated worm Response and Detection ImmuNe ALgorithm(CARDINAL)

As described in the previous section, different differentiation statuses of T-cells play varying roles in evoking overall immunity and tolerance in the HIS. This section introduces the overall architecture and components of the AIS that adopts CARDINAL, which employs various the T-cell immunity and tolerance mechanims reviewed in the previous section.

### 4.1 Overall Architecture

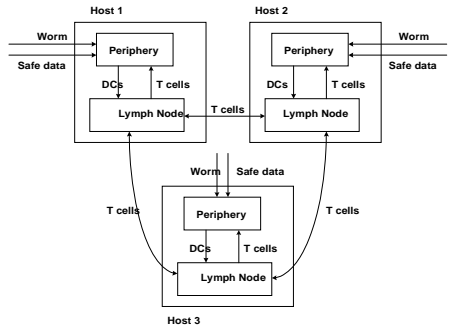

Fig. 1. Overview Architecture of CARDINAL.

The overall architecture of CARDINAL is presented in Fig. 1. It consists of periphery and lymph node processes [15]. Both processes reside on a monitoring host and any host running these two processes becomes a part of an artificial body which CARDINAL monitors. The periphery is comprised of DCs and various types of artificial T cells and they directly interact with input data such as network packets, email outbox or TCP connection requests etc. The input data also exists as a part of the periphery. DCs gather and analyse the input data and carry their analysis results to the lymph node. At the lymph node, naive T cells are created which subsequently differentiate into various types of effector T cells based on the input data analysis results continuously passed from DCs. Within CARDINAL, effector T cells are automated responders that react to worm related processes in the periphery. Effector T cells are assigned to a response target, a response type, and the number of peer hosts polled. Before the effector T cells migrate from the lymph node to the periphery, they interact with other effector T cells passed from peer hosts. This interaction allows locally generated effector T cells to determine whether they should perform assigned types of responses or not, and the numbers of peer hosts to be polled if they decide a response is appropriate. The local effector T cells assigned to particular responses, and the number of peer hosts to be polled are passed to the periphery processes at the local host and the peer hosts. These effector T cells now respond to the response targets, which are also defined as a part of the periphery process. In the next section, we provide more detailed descriptions of artificial cell interactions occuring at the periphery and lymph node processes within CARDINAL.

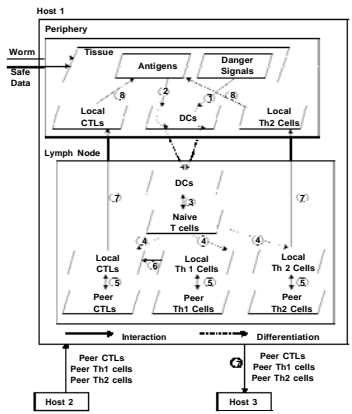
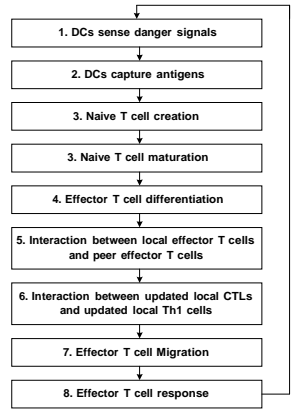

Fig. 2. Periphery and Lymph Node Processes in CARDINAL

Fig. 3. A flow chart of CARDINAL

4.2 Periphery and Lymph Node Processes

**DCs sense danger signals and capture antigens** The artificial tissue layer provides the location for two primary activities, the monitoring of danger signals (see ① in Fig. 2) and the collection of antigen in the form of input data (see ② in Fig. 2). Here danger signals are seen in the context of the symptoms arising from a worms infection. Well known worm infection symptoms include excessive cpu load at the host level, bandwidth saturation at the network level, and abnormal rates of email communication etc. Mechanisms of converting infection symptoms into danger signals that can be acted upon can be seen in [6] and are not discussed here. The DC's within CARDINAL then assess these danger signals and ascertain the severity of the attack and the certainty of its detection. The second purpose of the tissue layer is to provide a mechanism for the DC's to gain access to the input data reflecting the antigens, so that the threat level derived from the danger signal can be associated with its respective source and remembered. The extraction of antigen from the tissues by DC's is discussed in [4] [6].

**Naive T cell creation** Once collected in the periphery, DCs carry the danger signal assessment results and captured antigens to the lymph node. At the lymph node, naive T cells are created and these are subsequently differentiated based on the danger signal assessment results into their various states (see ❸ in Fig. 2). In nature, the receptors on naive T cells (TCR's) allow the cell to identify a particular type of antigen. For the sake of simplicity, our model assumes that the system will target the worm which always has a consistent attack signature and so can be detected by our naive T cells via these receptors. This assumption will be changed in future work to allow for the detection of polymorphic worms, which constantly change their form or functionality. In this way, the receptors of the naive T cells are simply copies of the antigens presented by DCs.

**Naive T cell maturation** Naive T cells continuously encounter DCs passed from the periphery (see ❸ in Fig. 2). During this process, DCs present danger signal assessment results to the naive T cells in three forms, as a form of a costimulatory signal and as two types of cytokines that reflect the potential danger signal, and each is affected differently based on the scale of the attack. The costimulatory signal is increased if a DC detects a severe attack, needing a strong response, and the certainty of that attack is assessed to be high. The cytokine IL-12 increases when a DC detects a severe attack requiring a strong response but with a relatively lower certainty, whereas the value of the cytokine IL-4 is incremented when a DC detects a less severe attack which only needs a weak response[2].

Naive T cells have three numerical values associated with them, these represent the "accumulated" certainties and severities of attacks recognised for each cell type: CTL activation values, Th1 activation values, and Th2 activation values. Whenever naive T cells interact with DCs, they evaluate whether the antigen presented by DCs are identical to their TCRs. If they are identical, naive T cells adjust these three activation values by taking account into the values of the costimulatory signals and the cytokines IL-12 and IL-4 produced by the DC's (see ❸ in Fig. 2). The costimulatory signal will influence the CTL activation value whilst IL-12 and IL-4 will influence the Th1 and Th2 activation values respectively. After a suitable period of time, these naive T cells are considered as ready to respond and differentiate.

**Effector T Cell differentiation** There are three different types of local effector T cells : local CTL, local Th1, and local Th2 cells (see ❹ in Fig. 2). The CTL activation , Th1 activation and Th2 activation values associated with the naive

---

[2] For a less severe attack, CARDINAL does not take into account the certainty of this kind of attack since a negative effect of a response triggered by a false positive error would be minor.

T cells will determine the types of local effector T cells that naive T cells will differentiate into. When one of these activation values exceeds a given threshold, via stimulation from the costimulatory molecules or cytokines from DCs, naive T cells will differentiate into the respective type of cell for which the threshold was exceeded. The newly differentiated local effector T cell will have an identical TCR pattern to the orginal naive T cell. In addition, they are cloned, and the number of clones reflects the numbers of polled peer hosts. This clonal rate is determined by the CTL, the Th1, and the Th2 activation values respectively. The larger the CTL activation value, the larger is the number of clones allocated to that CTL. Similarly, the larger the Th1 or Th2 cell activation values, the larger is the numbers of clones assigned to the Th1 cell or Th2 cell.

Interaction between local effector cells and peer effector cells Each type of local effector T cell only interacts with the same corresponding type of peer effector T cell transferred from the peer hosts (see ✪ in Fig. 2). This interaction takes place over four distinct stages. During the initial stage, at each host, CARDINAL selects local effector T cells whose numbers of clones are large enough to indicate that the antigens recognised by those effectors are severe in terms of their attack, and that the evidence of this attack is certain. During the second stage, CARDINAL reviews the local effector T cells that were not selected during the first stage and compares them to the peer effector T cells. Local effector T cells are then chosen if they match the required number of peer effector T cells, which detect the same antigens recognisied by local effector T cells. During the third stage, CARDINAL recalculates the number of clones assigned to the local effector T cells that were selected during stages one and two. The numbers of clones produced is determined by comparing the historical growth rate of the worm infection against the historical effector cell clone growth rate[3]. If the worm infection growth rate exceeds, or is equal to, the clone growth rate, CARDINAL increases the numbers of clones currently assigned to local effector T cells, otherwise CARDINAL decreases the numbers of clones of local effector T cells.

During the fourth and final stage, CARDINAL reviews the peer effector T cells received by the local host and identifies those cells that do not have a local effector T cell that are capable of detecting the same antigen. The numbers of clones assigned to these peer effector T cells is then decreased because those antigen have not been detected at this local host, and so are not considered a threat. Therefore, CARDINAL starts to suppress the response to that antigen. After this suppression, CARDINAL examines the the number of clones assigned to the peer effector T cells sent to the local host. If the number of clones exceeds zero, then this reflects a potential threat that the local host has yet to experience. In order to prepare the local host for this potential threat the local host will create a local naive T cell that is an exact copy of a peer effector T cell. This naive cell will have lower activation thresholds for its CTL, Th1 and Th2 activation values to ensure a rapid response is initiated to any subsequent antigen exposure. In this way, we create a form of memory within the CARDINAL system.

---

[3] The worm infection growth rate is estimated from the total number of responses which the peer hosts made during the previous two time steps. The clone growth rate is also measured as the change in the number of clones over the previous two time steps.

**Interaction between updated local CTLs and updated local Th1 cells**
Up to this point, effector T cells have only interacted with other effector T cells of the same type. However, CARDINAL also incorporates interactions amongst different types of effector cells. Before local effector T cells migrate to the periphery, another interaction between local CTLs and local Th1 cells occurs at the lymph node. During this interaction, the local Th1 cells can further increase the number of clones assigned to local CTL's if the two cells recognise the same antigen (see ⑥ in Fig. 2). As the certainty of an attack detected by a local Th1 cell is lower compared to that detected by a CTL, some fraction of the number of clones which a local Th1 cell has could be added to the number of clones of the local CTL. This variation in attack certainty between CTL's and Th1's depends on the type and timing of the danger signals' occurrence (infection symptoms). The interaction between a local Th1 and a local CTL would result in the fusion of various information related to an antigen, which is collected from diverse input sources over different time steps. This additional support from a Th1 cell reinforces the response of a CTL by increasing the number of CTL clones specific to that antigen. This is because they provide additional evidence as to the existence of an antigen threat.

**Effector T cell migration and response** After the cell interaction phase is complete, local and peer effector T cells with positive clone values begin a migration process either to respond to a threat in the periphery at a local level (see ⑦ ⑧ in Fig. 2) or communicate the existence of such a threat to other peer hosts (see ⑦ in Fig. 2). Local CTLs and local Th2 cells migrate to the periphery of the local host and commence their assigned response roles to counter the antigen attack. Th1 cells influence the number of CTL clones whilst in the lymph node, so their impact on the periphery is indirect. If the numbers of clones assigned to local effectors are positive, and there are no matching peer effector cells detecting identical antigens, CARDINAL creates new peer effectors which are copies of the local effectors. These new peer effector T cells, along with the existing peer effector T cells, migrate to other peer hosts if the number of clones associated with these cells is positive. This ensures that the knowledge of the antigen attack is communicated to the selected peer hosts. As described previously, the number of peer hosts selected for migration is determined by the severity and certainty of an attack. The actual hosts chosen for this migration subset are selected randomly from "all the possible peer hosts".

### 4.3 T cell Immunity and Tolerance within CARDINAL

As illustrated in previous sections, CARDINAL adopts various immune inspired components in order to implement an effective cooperative strategy for worm detection and response. Table 2 summarises these components and their roles within CARDINAL. In section 2, we highlighted three properties desirable for an effective worm detection and response system. We believe that CARDINAL would provide these properties through implementing T cell immunity and tolerance as follows:

– **Types of system responses should be determined by attack severity and certainty**: CARDINAL determines appropriate types of responses based on the attack severity and certainty assessed by DCs. DCs exposed to various types of danger signals produce different levels of costimulatory signals and cytokines, which in turn stimulate naive T cells recognising the antigen presented by DCs. The different degrees of accumulated costimulatory signals and cytokines reflect the severity and certainty of an attack measured collectively over multiple time steps and data sources. This kind of collective measurement would provide more accurate grounds to determine appropriate types of responses.
– **For performing adequate magnitudes of reponses, both local and peer information needs to be taken into account**: a local effector T cell assigned to a specific type of response can be further stimulated or suppressed by the interaction with peer effector T cells. This stimulation and suppression is realised through updating the number of clones assigned to each effector T cell, which performs a specific type of response.
– **Optimise the number of peer hosts polled**: CARDINAL optimises the number of clones(=the number of peer hosts polled) assigned to each effector T cell by dynamically estimating the severity of the worm's threat, the certainty of attack detection, and the growth rate of the infection. This estimation is implemented via several stages of different types of cell interactions. These interactions include tissue and DC, DC and naive T cell, local effector T cell and peer effector T cell, and local CTL and local Th1 cell interactions. As a result of these interactions, if CARDINAL considers the identified attacks to be more severe, certain, and to propagate faster, CARDINAL triggers a larger number of hosts to evoke an automated reponse. In addition, CARDINAL immediately suppresses the number of peer hosts polled when it observes that the severity and certainty of an attack becomes less, and the propagation speed of an observed attack becomes slower.

| CARDINAL Components | Roles | CARDINAL Components | Roles |
|---|---|---|---|
| Periphery | Input data access and reponding targets | Lymph Node | T cell creation, differentiation and interaction |
| Tissue | Local anomaly detectors | DC Costimulatory Signals | Frequencies of severe and certain attacks |
| DC Cytokine IL12 | Frequencies of severe and less certain attacks | DC Cytokine IL4 | Frequencies of less severe attacks |
| Danger Signals | Infection symptoms | Antigens | Attack Signatures |
| TCRs | Attack signatures | CTLs | Strong Automated Responders |
| Th1 Cells | CTL controller | Th2 Cells | Weak Automated Responders |
| Activation Values of a Naive T cell | Accumulated severities and certainties of attacks | Number of clones of an Effector T cell | Number of polling peer hosts |

Table 2. CARDINAL components and their roles

The current mechanisms within CARDINAL, inspired by T cell immunity and tolerance, would provide these three desirable properties, which will help an automated worm detection and response system to reduce a false positive error.

## 5  Conclusion

In this paper, we have shown how the link between the the innate immune system(DCs) and the adaptive immune systems(T-cells), can be computationally modelled to form the basis of a novel worm detection algorithm. In particular, we identified three key properties of T- cell and mapped these into the CARDINAL system: *T-cell proliferation - to optimise the number of peer hosts polled. *T-cell differentiation - to assess attack severity and certainty and *T-cell modulation and interaction - to balance local and peer information.

Further extensions of the presented T-cell algorithm are possible. In particular, performance could be enhanced by including the notion of antigen generalisation leading to T-cell memory. Additionally, immunologists have recently discovered a potentially third T-cell line in the shape of regulatory T-cells. It is currently thought that these cells form an important part in inducing tolerance by regulating other T-cell behaviour. However, more details have yet to emerge before this class of cell can be efficiently incorporated into our computational model.

It is also worth noting here that the proposed T-cell algorithm does not operate in isolation, but in unison as a part of the novel danger theory inspired system [1]. Thus, it is essential for the algorithm to work with artificial tissue [4] and dendritic cell algorithms [6]. Once integrated, these systems should mirror the robustness and effectiveness of their human counterparts.

Current work is focusing on implementing a simulated model of AIS adopting CARDINAL. To reflect worm propagation in the real world, the simulated model needs to accommodate a number of settings and parameters such as the type of worm (random-scan worm or topology-based worm), a network topology, a rate of worm infection depending on selected worm types and the network topology etc. In order to provide such a realistic environment in the CARDINAL simulated model, the epidemic models defining the state transitions and conditions of infections are being currently studied [3] [13] [14].

## Acknowledgements

This project is supported by the EPSRC (GR/S47809/01), Hewlett-Packard Labs, Bristol, and the Firestorm intrusion detection system team. Special thanks to Jamie Twycross for initiating the study of worm detection problems. Great thanks to all the members of the "Danger Project" (www.dangertheory.com) for their helpful feedback and inspiring discussion.